\pdfoutput=1

\documentclass[11pt]{article}
\usepackage{times,latexsym}
\usepackage{url}
\usepackage[T1]{fontenc}

\usepackage[]{acl}

\usepackage{xspace,mfirstuc,tabulary}

\usepackage{graphicx}
\usepackage{amsfonts}
\usepackage{amsmath}
\usepackage{url}
\usepackage{mathtools}
\usepackage{mathdots}
\usepackage{color}
\usepackage{siunitx}
\usepackage{array}
\usepackage{multirow}
\usepackage{tabularx}
\usepackage{booktabs}
\usepackage{cleveref}
\usepackage{xcolor}
\usepackage{makecell}
\usepackage{microtype}
\usepackage[normalem]{ulem}  
\useunder{\uline}{\ul}{}

\title{SHARE: a System for Hierarchical Assistive Recipe Editing}

\author{Shuyang Li \\
  UC San Diego \\
  \texttt{lishuyang@meta.com} \\\And
  Yufei Li \\
  UC Riverside \\
  \texttt{yli927@ucr.edu} \\\AND
  Jianmo Ni \\
  UC San Diego \\
  \texttt{jianmon@google.com} \\\And
  Julian McAuley \\
  UC San Diego \\
  \texttt{jmcauley@ucsd.edu}
 }

\date{}

\begin{document}
\maketitle
\begin{abstract}
    The large population of home cooks with dietary restrictions is under-served by existing cooking resources and recipe generation models.
    To help them, we propose the task of \emph{controllable recipe editing}: adapt a base recipe to satisfy a user-specified dietary constraint.
    This task is challenging, and cannot be adequately solved with human-written ingredient substitution rules or existing end-to-end recipe generation models.
    We tackle this problem with SHARE: a \textbf{S}ystem for \textbf{H}ierarchical \textbf{A}ssistive \textbf{R}ecipe \textbf{E}diting, which performs simultaneous ingredient substitution before generating natural-language steps using the edited ingredients.
    By decoupling ingredient and step editing, our step generator can explicitly integrate the available ingredients.
    Experiments on the novel \emph{RecipePairs} dataset---83K pairs of similar recipes where each recipe satisfies one of seven dietary constraints---demonstrate that SHARE produces convincing, coherent recipes that are appropriate for a target dietary constraint.
    We further show through human evaluations and real-world cooking trials that recipes edited by SHARE can be easily followed by home cooks to create appealing dishes.
\end{abstract}

\begin{figure*}
  \includegraphics[width=\textwidth]{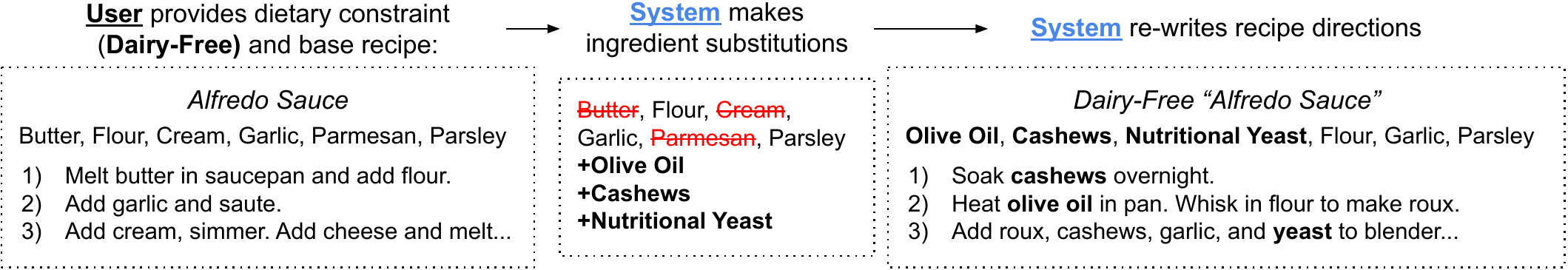}
  \caption{We investigate the task of controllable recipe editing: edit a base recipe to satisfy a given dietary restriction.}
  \label{fig:task}
\end{figure*}

\section{Introduction}

Cooking has played an integral role in human civilization and evolution for over 1.8 million years \cite{wrangham2009catching}.
A growing population follows some form of dietary restriction \cite{goldberg2017ific}, with motivations ranging from socioeconomic to medical \cite{shepherd1996constraints, 10.1001/jama.2015.18396}.
Home cooks browsing recipes on online recipe aggregators may
often encounter an interesting recipe that does not fit their dietary needs (e.g.~vegetarianism), and would benefit from a way to \emph{edit} that recipe to fit their needs.
These restrictions are easy for users to specify, but existing recipe websites offer few options for users with dietary constraints---even those with common restrictions like gluten intolerance (\Cref{tab:underserved}).
We see an opportunity to build a system to tailor recipes to fit users' preferences and dietary constraints.

To help such users, we introduce the task of \emph{controllable recipe editing}: editing a base recipe so that it satisfies a user-specified dietary constraint (\Cref{fig:task}).
Controllable recipe editing can help home cooks of any experience level find diverse ways to satisfy their dietary needs and help develop interactive, personalized products like meal kits to accommodate individual needs and preferences.
This is a challenging task: rule-based substitution methods and existing recipe generators cannot adequately account for both the dietary and structural impacts of ingredient substitutions.

To tackle this task, we propose a \textbf{S}ystem for \textbf{H}ierarchical \textbf{A}ssistive \textbf{R}ecipe \textbf{E}diting (SHARE).
We first use a Transformer \cite{DBLP:conf/nips/VaswaniSPUJGKP17} encoder-decoder to perform multiple simultaneous ingredient substitutions conditioned on a dietary restriction.
We next employ a Transformer language model to write new instructions conditioned on the edited ingredients, using a copy mechanism \cite{DBLP:conf/acl/GuLLL16} to increase ingredient-step coherence.

We conduct experiments on a novel dataset of 83K recipe pairs from
user-reviewed recipes on a popular recipe site.
Each pair is associated with a \emph{dietary restriction} and contains a \emph{base} recipe and a similar \emph{target} recipe that satisfies the constraint.
We evaluate edited recipes
via automatic metrics to demonstrate that SHARE produces diverse and high-fidelity recipes.
We survey 672 home cooks to assess the quality of our edited recipes compared to human-written recipes, finding that SHARE consistently produced the highest quality edits---in the process discovering several stylistic and structural aspects that cooks evaluate when searching for recipes.
We also recruit seven home cooks to cook 21 recipes generated from SHARE and evaluate both the cooking process and final product, showing that such recipes are delicious, easy to make, and satisfy dietary restrictions.

We summarize our main contributions:
1) We propose the \emph{controllable recipe editing} task to assist an under-served population of home cooks by editing recipes to satisfy their dietary constraints;
2) We create the novel \textbf{RecipePairs} dataset---containing 83K pairs of recipes and versions thereof satisfying a dietary constraint---to facilitate recipe editing;
3) We train a hierarchical model for controllable recipe editing (SHARE), finding via quantitative trials, human studies, and real-world cooking trials that SHARE creates more coherent and constraint-respecting recipes compared to strong baselines---and that home cooks find its dishes appealing.

\section{Related Work}

We specifically aim to help a population under-served by existing resources: people with dietary restrictions.
Recent works in nutritional recipe recommendation \cite{chen2019eating, gorbonos2018nutrec} aim to recommend generally healthier food options.
To our knowledge, while prior work has focused on ingredient substitution \emph{or} generating recipe directions, ours is the first to examine both to create a complete recipe.
We are motivated by work on single ingredient substitution \cite{DBLP:conf/kes/YamanishiSNFK15, DBLP:conf/websci/TengLA12}, but our system accommodates multiple simultaneous substitutions when predicting a target ingredient set.
Recent work in recipe generation has focused on improving coherence \cite{DBLP:conf/iclr/BosselutLHEFC18} and ingredient specificity \cite{DBLP:conf/emnlp/KiddonZC16}.
We draw from these as well as sentence editing work that uses retrieved \emph{prototype sequences} to control generated text \cite{DBLP:journals/tacl/GuuHOL18, DBLP:conf/nips/HashimotoGOL18}, to improve conditioning on ingredients.

The editing of procedural texts like recipes resembles conditional text generation and paraphrase generation.
Transformer \cite{DBLP:conf/nips/VaswaniSPUJGKP17} language models have seen success in both fields \cite{DBLP:journals/corr/abs-1909-05858, DBLP:conf/emnlp/WitteveenA19}, and \citet{DBLP:journals/corr/abs-1910-10683} demonstrated that such models can make use of a set of control codes for simultaneous multi-task learning.
\citet{DBLP:conf/www/LeeKAPLLV20} train a single model for two tasks---ingredient extraction and recipe step generation conditioned on the recipe directions and ingredients, respectively.
A user must thus provide either the complete set of ingredients or a full set of directions---neither of which may be known to home cooks looking for ways to adapt a recipe to their individual needs.
\citet{DBLP:conf/emnlp/MajumderLNM19} explore ways to use browsing histories to personalize recipes for cooking website users.
Their model uses these histories, a recipe name, and ingredients as input to generate personalized instructions, but cannot be controlled to generate constraint-satisfying (e.g.~gluten-free) recipes.

\section{RecipePairs Dataset}
\label{sec:data}

Several recipe corpora have been collected, including the 150K-recipe \emph{Now You're Cooking!} dataset \cite{DBLP:conf/emnlp/KiddonZC16,DBLP:conf/iclr/BosselutLHEFC18},
Recipe1M+ \cite{marin2019learning} for cross-modal retrieval tasks, and the Food.com \cite{DBLP:conf/emnlp/MajumderLNM19} dataset.
We extend the Food.com dataset, aggregating user-provided category tags for each of the 459K recipes---comprising soft constraints (low-sugar, low-fat, low-carb, low-calorie) and strict/hard dietary restrictions (gluten-free, dairy-free, vegetarian).
We compile a list of 501 ingredients that violate hard constraints following medical website guidance,\footnote{e.g.~\url{https://www.ncbi.nlm.nih.gov/books/NBK310258/}} further filtering out any tagged recipes that contain banned ingredients to ensure each of our category tags is accurate.

We pair recipes into a \emph{base} recipe that does not satisfy any \emph{category} (dietary constraint) and a \emph{target} version that satisfies the constraint.
To ensure base and target recipes are similar, we pair recipes by name, sampling target recipes whose name contains a base recipe name (e.g.~`healthy oat chocolate cookie' as a gluten-free version of `chocolate cookie').
We also filtered out pairs of recipes with low ingredient overlap, as well as manually verifying a subset of 20\% of recipe pairs to ensure quality.
This process resulted in 83K total pairs in RecipePairs comprising 36K unique base recipes and 60K unique targets.

We split the data into 81K pairs for training and 1K disjoint pairs each for testing and validation.
The large ingredient space contributes to the difficulty of the task.
We normalize ingredients by stripping brand names and amounts (e.g.~`Knoxville Farms' or `1 tablespoon') and remove rare variants, resulting in 2,942 unique ingredients that appear across 10+ recipes in our dataset.
In \Cref{tab:underserved}, we show the number of recipe pairs targeting each dietary constraint and the number of banned ingredients for each hard constraint.

\paragraph{Recipes Satisfying Dietary Restrictions}
To better understand the audience that would benefit from a recipe editing application, we first investigate how well users of Food.com are served when searching for recipes satisfying dietary constraints.
A long tail of dietary constraints are poorly supported by the site.
While 93\% of users look for recipes that satisfy some dietary constraint, only 43\% of all recipes satisfy a dietary restriction---the stark difference between the proportion of users looking to follow a dietary constraint and the proportion of recipes accommodating each constraint is seen in \Cref{tab:underserved}.
In particular, low-sugar, dairy-free, and gluten-free recipes are lacking: collectively they make up less than 10\% of available recipes despite one fifth of users looking for such recipes.
Up to 7\% of the American population suffers from a form of gluten allergy
and up to 20\% opt for a gluten-free diet \cite{igbinedion2017non}, but less than 2.5\% of recipes on Food.com are labeled as gluten-free.
Users looking for recipes that fit their requirements may thus be discouraged from using recipe sites.

\section{Task}
\label{sec:task}
To help home cooks discover delicious and appropriate recipes, we desire a system for \emph{controllable recipe editing}.
A user should be able to specify the \emph{base} recipe they would like to edit and the constraint to satisfy.
Our model should output a similar but novel recipe that satisfies the constraint.
To perform editing on the RecipePairs corpus, we consider a \emph{base} recipe $\mathcal{R}_b=\{N_b; I_b; S_b\}$ comprising name $N_b$, ingredients $I_b$, and directions $S_b$.
This recipe does not satisfy a dietary constraint $c\in \mathcal{C}$ (e.g.~low-fat, dairy-free).
The goal of our model is to transform the base recipe into some \emph{target} recipe $\mathcal{R}_t=\{N_t; I_t; S_t\}$ that satisfies $c$.
In this paper we investigate editing recipes to satisfy a single specified dietary constraint; in future work we aim to extend our method to accommodate multiple restrictions.
In our System for Hierarchical Assistive Recipe Editing (\textbf{SHARE}), we break the task into two sequential parts: 1) predicting target ingredients $I_t$ given the base name $N_b$ and ingredients $I_b$ alongside dietary constraint $c$, and 2) generating recipe directions $S_t$ from the new ingredients $\hat{I}_t$.

\begin{table}[t!]
\small
\centering
\begin{tabular}{@{}lrrrrr@{}}
\toprule
            & \hspace{-10mm} \% Users & \hspace{-3mm} \% Recipes & \hspace{-1mm} Pairs & Rules & Banned \\ \midrule
Low-Carb    & 71.3     & 16.9       & 19K   & 10    & --     \\
Low-Calorie & 63.8     & 14.3       & 17K   & 78    & --     \\
Low-Fat     & 51.5     & 8.6        & 13K   & 55    & --     \\
Low-Sugar   & 21.9     & 2.6        & 6K    & 11    & --     \\ \midrule
Vegetarian  & 59.9     & 13.3       & 18K   & 83    & 252    \\
Gluten-Free & 24.1     & 2.3        & 6K    & 18    & 137    \\
Dairy-Free  & 19.7     & 1.5        & 4K    & 14    & 112    \\ \bottomrule 
\end{tabular}
\caption{For each soft (top) and hard (lower) dietary constraint: 
users searching for such recipes, number of recipes satisfying constraint, pairs in RecipePairs, ingredient substitution rules, and prohibited ingredients.}
\label{tab:underserved}
\end{table}

\paragraph{Why isn't substitution enough?}
At first glance, it may seem acceptable to simply replace problematic ingredients with fixed substitutes.
To test this, we compiled 269 dietary substitution rules (\Cref{tab:underserved}) from medical websites\footnote{e.g. \url{https://www.mayoclinic.org/}} into the \textbf{Rule} baseline that replaces all constraint-violating ingredients in a recipe
(e.g.~for dairy-free, butter to margarine).
In the directions, we replace references to removed ingredients with the substitutes (e.g.~beat sugar and butter $\rightarrow$ beat sugar and \emph{margarine}).
For constraint-violating ingredients that do not have a documented substitution, this model removes the ingredient and all references in the text.
This method struggles to predict target ingredients (\Cref{tab:ingr-auto-eval}), and cannot account for subtle changes (e.g.~cooking techniques) necessary to accommodate new ingredients.
Indeed, while \emph{nutritional} impacts of ingredient substitutions are easily inferred, they have intricate effects on recipe structure and palatability.
This suggests that recipe editing is a challenging task that cannot be solved by simple rule-based systems.

\begin{figure*}
  \includegraphics[width=0.95\linewidth]{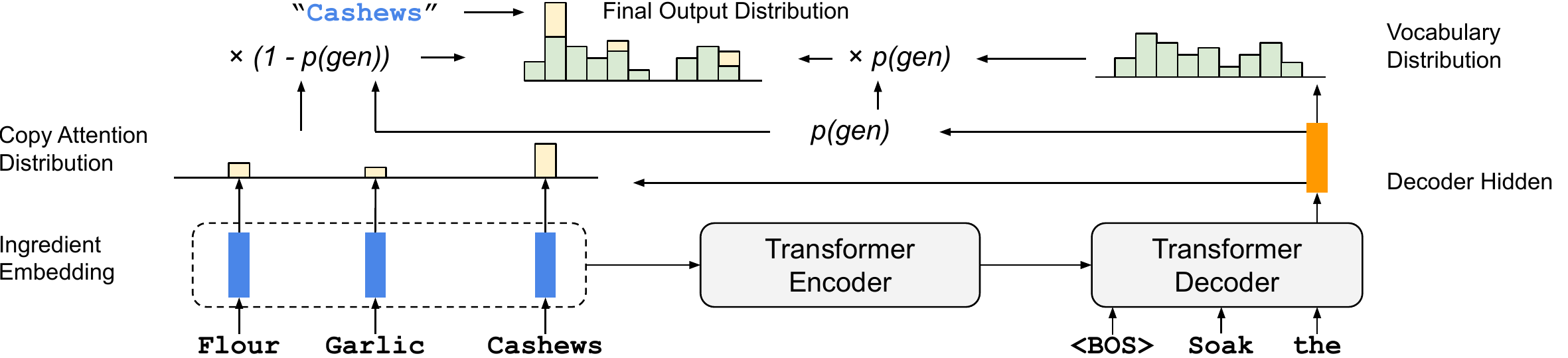}
  \caption{
  SHARE step generation module.
  The final hidden state at each position is used to calculate  $p_\text{gen}$, weighing the probability of copying tokens directly from the input ingredients vs. generating from the vocabulary distribution.}
  \label{fig:step-model}
\end{figure*}

\section{SHARE: a Hierarchical Edit Model}
\label{approach}
We pose \emph{controllable recipe editing} as a supervised learning task: predict a target recipe conditioned on a base recipe and a provided constraint.
Our SHARE model consists of two independently trained components for ingredient editing (\Cref{sec:ingr-model}) and step prediction (\Cref{sec:step-model}).
As ingredient quantities in a recipe may vary based on serving size, we omit exact amounts when editing ingredient lists.
When asked to cook recipes edited by our system, home cooks find it relatively easy to decide the amount of each ingredient to use (\Cref{human}).
To ensure that recipes produced by our systems are 100\% safe for cooks with dietary constraints, we 
can further filter edited ingredient lists and blacklist inappropriate ingredient tokens from being generated when using our system in the real world.
We discuss the instrumentation and impact of filtering and blacklisting modules in \Cref{satisfying_constraints}.

\subsection{Ingredient Editing}
\label{sec:ingr-model}

We pose ingredient editing as a multi-label binary classification task: learn the likelihood of an ingredient appearing in the target recipe, conditioned on the base recipe name and ingredients as well as a dietary constraint.
First, we embed the category ID and base recipe name via a $d$-dimensional learned embedding matrix and encode them with a $d$-dimensional Transformer \cite{DBLP:conf/nips/VaswaniSPUJGKP17} encoder.
We then pass the base recipe ingredient IDs into a $d$-dimensional Transformer decoder---at each position, each layer of the decoder applies attention over previous positions (self-attention) as well as the encoder outputs.
The decoder outputs are projected into an $|\mathcal{I}+1|$-dimensional vector representing logits for each ingredient in our vocabulary as well as the \emph{eos} token.

A typical transformer decoder predicts ingredients auto-regressively in an ordered list until the \emph{eos} token is encountered: $P(\hat{I}_{t,k}|R_b, c, \hat{I}_{t,<k})$.
This strategy penalizes for the order of ingredients, but the ingredients used in a recipe are an un-ordered set---\emph{butter and flour} is the same as \emph{flour and butter}.
To remove this order dependence when editing ingredients, we adopt the Set Transformer \cite{DBLP:conf/cvpr/SalvadorDNR19} strategy: we max-pool logits for each ingredient in our output sequence across positions:
\begin{equation}
    P(i\in \hat{I}_t|R_b, c)\ \ \forall\ \ i\in \mathcal{I}
\end{equation}
returning the $K$ ingredients with the highest score overall.
We ignore the \emph{eos} logit in the max-pooling; its position denotes the number of predicted ingredients $K$.

We train the model by maximizing binary cross-entropy (BCE) loss between the pooled ingredient logits and ground truth:
\begin{equation}
    \mathcal{L} = -\frac{1}{|\mathcal{I}|}\sum_{i \in \mathcal{I}} \mathbb{I}^{+} P(i|R_b, c)+\mathbb{I}^{-} (1 - P(i|R_b, c))
\end{equation}
where $\mathbb{I}^{+}$ and $\mathbb{I}^{-}$ indicate whether ingredient $i$ belongs to the true target recipe.

We also include a cardinality loss of dimensionality $T$: the BCE loss between the \emph{eos} logit at all time steps $t \leq T$ compared to the ground truth (1 at position $K$ and 0 otherwise).
At inference time, we predict $K$ via Bernoulli sampling from the \emph{eos} logit at each position.

\subsection{Step Generation}
\label{sec:step-model}

We next train a language model that takes our edited set of ingredients as input and generates a new set of recipe steps as a single string (\Cref{fig:step-model}).
This model has no explicit dependency on the base recipe, allowing us to train the model on all 459K recipes from our dataset (excluding evaluation pairs).
We follow a conditional language modeling setup: 

\begin{equation}
    P(S|\hat{I}_t) = \prod^K_{k=0} P(s_k|s_{k-1} \ldots s_0, \hat{I}_t)
\end{equation}

We represent the sequence of input ingredients $\hat{I}_t$ via their names, joined by a comma for a total of $K_\text{ingr}$ tokens.
Each token $w_i$ is embedded in $d$ dimensions to form the sequence $e_0, \ldots, e_{K_\text{ingr}}$, which is then encoded via a Transformer encoder.
We obtain a $d$-dimensional hidden state $h_k$ for each position $k$ in our directions sequence via a Transformer decoder that attends over previous positions as well as the encoder outputs.
These hidden states are projected into our vocabulary space $\mathcal{W}$ with a matrix tied to the input embeddings to produce a vocabulary distribution $P(s_k|\hat{I}_t, s_{<k})$ at each position.
While this encoder-decoder setup allows us to perform fluent language modeling, it can struggle to produce coherent recipes that properly make use of input ingredients \cite{DBLP:conf/emnlp/KiddonZC16}.
Thus, we propose to bias our model toward stronger ingredient conditioning by applying copy attention over our input ingredient embeddings.

\paragraph{Ingredient Copy Attention}
Our model can directly copy tokens from the input ingredients sequence via a copy attention mechanism \cite{DBLP:conf/acl/GuLLL16,DBLP:conf/acl/SeeLM17}.
At each position $k$ of our directions, we calculate scaled dot-product attention \cite{DBLP:conf/nips/VaswaniSPUJGKP17} weights $\alpha^i_k$ over input ingredient tokens $i$ using the final decoder hidden state as query \textbf{Q}
and a concatenation of our learned input token embeddings as key
\textbf{K}:
\begin{equation}
    \begin{aligned}
    \alpha^i_k &= \text{softmax}(\frac{\textbf{Q}\textbf{K}^\text{T}}{\sqrt{K_\text{ingr}}})\\
    \textbf{Q} &= h_k \in \mathbb{R}^{1,d}\\
    \textbf{K} &= [ e_0; \ldots; e_{K_\text{ingr}} ] \in \mathbb{R}^{K_\text{ingr}, d}
    \end{aligned}
\end{equation}

We thus learn a distribution
representing a chance of copying an ingredient token directly from the input sequence.

At each time-step, our model calculates a generation probability $p_\text{gen}\in[0,1]$ via a learned linear projection from the decoder output $h_k$:
$p_\text{gen} = \sigma(W_\text{gen}^\text{T} h_k + b_\text{gen})$
where $\sigma$ is the sigmoid function.
We use $p_\text{gen}$ as a soft gate to choose between generating the next token from the vocabulary distribution or copying a token from the input ingredients, giving us the final output distribution:
\begin{equation}
    P(w) = p_\text{gen} P_\text{vocab}(w) + (1 - p_\text{gen}) \sum_{i:w_i=w} \alpha^i_k
\end{equation}

At training time, we compute cross-entropy loss over the predicted recipe directions against the target recipe steps, minimizing the log likelihood of the predicted sequence via the factorized joint distribution:
$P(S|\hat{I}_t)=\prod^n_j P(s_{k}|s_{<k}, \hat{I}_t)$
using the final output distribution.

\section{Experimental Setup}
\label{exp-setup}

\paragraph{Baselines}

While controllable recipe editing has not been attempted to our knowledge, we adapt state-of-the-art recipe generation models as strong baselines alongside the substitution rule method (\textbf{Rule}) described in \Cref{sec:task}.
\citet{DBLP:conf/www/LeeKAPLLV20} fine-tuned a large language model (LM)
on Recipe1M+
to extract ingredients from recipe directions and write directions given ingredients: RecipeGPT.

We adapt RecipeGPT for multi-task editing
(\textbf{MT}):
1) predict target ingredients from the category, base recipe name, and base recipe steps by modeling the concatenated sequence $[c; N_b; S_b; I_t]$; and 2) generate target steps from the category, base recipe name, and target recipe ingredients by modeling the concatenated sequence $[c; N_b; I_t; S_t]$.

We next
fine-tune RecipeGPT for end-to-end (\textbf{E2E}) recipe editing by priming the model \cite{DBLP:journals/corr/abs-1909-05858} with a dietary constraint and base recipe name $[c; N_b]$ (e.g.~`low\_fat cheesecake') to generate target ingredients and steps in a single sequence.
E2E is twice as efficient as MT and simultaneously predicts ingredients and steps.

We also investigate a form of prototype editing in our end-to-end model, by treating the full base recipe as a prototype which needs to be edited to accommodate a constraint (\textbf{E2E-P}).
While methods for sentence editing use a latent edit vector in addition to a prototype sentence \cite{DBLP:journals/tacl/GuuHOL18}, we prime our model with the fully specified base recipe ($[c; N_b; I_b; S_b]$) in addition to the category to predict ingredients and steps of an edited recipe.

\paragraph{Evaluation Metrics}
We first compare SHARE to baseline models via quantitative experiments on RecipePairs.
To measure \emph{fidelity} of edited ingredient lists against ground truth, we calculate Jaccard similarity (IoU) and F1 scores \cite{DBLP:conf/cvpr/SalvadorDNR19}, and precision and F1 for insertions and deletions.
To measure \emph{step fidelity}, we compare the workflow of two recipes via Normalized Tree Edit distance (NTED) between structured ingredient-cooking-action tree representations \cite{DBLP:conf/chi/ChangGJHKA18}.
These trees have ingredients as root nodes, with intermediate nodes drawn from a curated set of 259 lemmatised cooking action verbs.

We measure the \emph{fluency} of generated directions via ROUGE-L \cite{rouge}, a metric that assesses \emph{n}-gram overlap between edited and gold recipe texts.
We measure \emph{diversity} via the percentage of distinct bigrams (D-2) across edited recipes.
To assess whether edited recipes are \emph{appropriate} given the target dietary constraint, we extract all ingredient mentions from recipe directions to flag if ingredients in the edited ingredients list and directions violate the target constraint.
We also conduct a human study to critique generated recipes from each model and real-world cooking trials where home cooks follow recipes generated by SHARE to produce dishes (\Cref{human}).

\section{Results}
\label{results}

\paragraph{RQ 1: How accurately does SHARE edit ingredient lists?}
In \Cref{tab:ingr-auto-eval} we compare edited ingredient lists to ground truth.
Ingredient substitution is a challenging task---human-written substitution rules lack the coverage and flexibility required to accurately edit recipes to satisfy dietary constraints.
SHARE outperforms all baselines in overall editing accuracy (IoU and F1).
We find that correctly adding ingredients is significantly more challenging than removing inappropriate ingredients, with our approach achieving the highest precision in both cases.
Baseline LM methods (MT, E2E, E2E-P) predict new ingredients from the vocabulary of English language tokens.
SHARE instead separates the ingredient- and step-editing tasks in the hierarchical pipeline, predicting ingredient sets from a known ingredient space.
This approach allows us to learn how specific ingredients interact in context of a recipe and dietary restrictions.

\begin{table}[t!]
\centering
\small
\begin{tabular}{@{}lll|ll|ll@{}}
\toprule
 & \multicolumn{1}{l}{} & \multicolumn{1}{l|}{} & \multicolumn{2}{c|}{Insertion} & \multicolumn{2}{c}{Deletion} \\ \midrule
Model & IoU  & F1   & F1   & Prec. & F1   & Prec. \\ \midrule
Rule  & 22.2 & 33.9 & 1.2  & 2.1       & 18.4 & 42.0      \\
MT    & 31.6 & 45.8 & 25.6 & 28.9      & 69.5 & 82.4      \\
E2E   & 29.5 & 43.2 & 21.1 & 26.8      & 69.5 & 79.9      \\
E2E-P & 30.6 & 44.7 & 26.2 & 29.5      & \textbf{73.2} & 81.0      \\
\textbf{SHARE}  & \textbf{33.0*} & \textbf{47.5*} & \textbf{26.7} & \textbf{35.2*}      & 66.1 & \textbf{83.2}      \\ \bottomrule
\end{tabular}
\caption{Fidelity of edited ingredient lists when compared to target recipe ingredients, in terms of overall IoU/F1 and insertion/deletion F1 and precision. SHARE creates significantly ($p < 0.05$) higher fidelity edits compared to baselines.}
\label{tab:ingr-auto-eval}
\end{table}

\begin{table}[t!]
\small
\centering
\begin{tabular}{@{}lllr@{}}
\toprule
              & ROUGE $\uparrow$ & NTED $\downarrow$                 & D-2 $\uparrow$ \\ \midrule
Rule          & 20.3 $\pm$ 11.5              & 0.623 $\pm$ 0.072            & {\ul 24.48}            \\
MT            & \textbf{23.3} $\pm$ \textbf{9.1}      & 0.618 $\pm$ 0.073            & 7.28                   \\
E2E           & 21.2 $\pm$ 8.9               & 0.621 $\pm$ 0.074            & 10.13                  \\
E2E-P         & 23.0 $\pm$ 9.1               & 0.621 $\pm$ 0.068            & 7.67                   \\ \midrule
SHARE         & 22.6 $\pm$ 7.4               & \textbf{0.611} $\pm$ \textbf{0.065*} & \textbf{13.04}         \\
\ \ \ Paired Data   & 22.2 $\pm$ 7.2               & 0.614 $\pm$ 0.063            & 13.13                  \\
\ \ \ No Copy Attn. & 20.2 $\pm$ 7.5               & 0.622 $\pm$ 0.065            & 12.04                  \\ \bottomrule
\end{tabular}
\caption{Fluency (ROUGE), fidelity (NTED), and diversity (D-2 \%) metrics for edited recipe directions. SHARE creates significantly ($p < 0.05$) more structurally similar recipes to the target.}
\label{tab:step-auto-eval}
\end{table}

\paragraph{RQ 2: How reasonable are the recipe steps generated by SHARE?}
We measure recipe direction quality in terms of fluency, diversity, and fidelity (\Cref{tab:step-auto-eval}).
Here again, the Rule baseline's tendency to simply remove ingredient references not found in human-written substitution rules hurts recipe coherence.
Despite other baselines leveraging large pre-trained LMs, no model reports statistically significantly better ROUGE scores than all alternatives.
We confirm observations from \citet{DBLP:conf/emnlp/BahetiRLD18} and \citet{DBLP:conf/emnlp/MajumderLNM19} that \emph{n}-gram metrics like ROUGE correlate poorly with human judgment (\Cref{human}).
Our hierarchical ingredient-to-step generation format explicitly decouples the base recipe and edited instructions, allowing SHARE to create more diverse recipes than large LM baselines (+3-5\% bigram diversity).
We do not require paired data and can train SHARE on all 400K+ recipes from Food.com; we find that a variant trained only on the same paired corpus (\textbf{Paired Data}) still out-performs baselines for diversity and fidelity.
Thus, independent of corpus size, our hierarchical approach allows our step generator to learn 
that many dishes can be cooked from the same ingredients---reflecting
the many satisfactory ways of editing a recipe to fit a dietary constrain---and addresses a flaw we observe with existing recipe aggregators: a lack of diverse recipes satisfying dietary constraints.

SHARE generates recipes that are not only diverse but also high-fidelity: when comparing workflows between generated recipes and ground truth it significantly out-performs all baselines ($p < 0.05$),
producing recipes that are the most structurally similar to gold (lowest average NTED).
This reflects how our hierarchical step generator is trained to generate recipes conditioned solely on input ingredients and suggests that it best learns how human cooks make use of a given set of ingredients.
We also confirm the importance of our ingredient copy attention module, as a pure Transformer encoder-decoder (\textbf{No Copy Attn.}) edits recipes with significantly worse structural fidelity.

\paragraph{RQ 3: How well does SHARE satisfy dietary constraints?}
\label{satisfying_constraints}
Previous work in recipe generation has focused on writing recipe steps; we aim to instead create complete recipes satisfying a user's dietary constraints.
As some human-written recipes satisfying a soft constraint (e.g.~low-sugar) nonetheless judiciously use unhealthy ingredients (e.g.~corn syrup), we analyze strict dietary constraints: dairy-free, gluten-free, and vegetarian recipes.
Violating such constraints can often result in health risks,
so we track the percentage of edited recipes that use prohibited ingredients (from \Cref{sec:data}).

SHARE makes significantly more appropriate edits to ingredient lists compared to alternatives (\Cref{tab:cat-err-ingr}), with less than 3\% of edited recipes containing a prohibited ingredient in its ingredients list.
Recipe directions, however, are generated as unstructured text by SHARE and LM baselines and can reference problematic ingredients not in the ingredients list.
As such, we identify ingredient references in generated steps, and find a 4.5-6.5\% violation rate (\Cref{tab:cat-err-step}).

\begin{table}[t!]
\centering
\small
\begin{tabular}{@{}lrrrr@{}}
\toprule
Model & Dairy    & Gluten   & Vegetarian    & Overall       \\ \midrule
Rule  & \textit{0.00}         & \textit{0.00}         & \textit{0.00}         & \textit{0.00}         \\ \midrule
MT    & 20.55         & 32.89         & 10.24         & 17.23         \\
E2E   & 39.73         & 46.05         & 21.46         & 30.51         \\
E2E-P & 10.96         & 34.21         & 4.88          & 12.43         \\
\textbf{SHARE}  & \textbf{4.11} & \textbf{9.21} & \textbf{0.00} & \textbf{2.82} \\ \bottomrule
\end{tabular}
\caption{Percent of edited ingredients lists violating target hard dietary restrictions (before filtering).}
\label{tab:cat-err-ingr}
\end{table}

Even if SHARE generates appropriate recipes >95\% of the time, a 4.5\% chance of generating a potential dangerous recipe edit remains unacceptable for production use.
Filtering out prohibited ingredients from the ingredients list helps reduce the incidence of bad references for SHARE, while it has no impact on other LM baselines.
This further suggests our approach conditions more strongly on input ingredients when writing steps compared to RecipeGPT-based methods.
We can further remove all problematic ingredient references by setting the likelihood of prohibited ingredient sequences to zero during step generation.

\paragraph{Performance vs. Efficiency}
In addition to out-performing baseline models for recipe editing, SHARE is much smaller than such models, with 12.5M learnable parameters compared with 124M (\textasciitilde10x larger) for RecipeGPT-based models.
Its smaller size confers additional benefits: it takes on average 3.9s to generate a recipe from E2E-P and MT, while it takes on average only 0.9s (\textbf{4.3x faster}) for SHARE.
Thus, a hierarchical approach to controllable recipe editing accommodates the training of lightweight models for each sub-task that can serve users with acceptable latency.

\begin{table}[t!]
\centering
\small
\begin{tabular}{@{}lrrrr@{}}
\toprule
Model      & Dairy          & Gluten         & Vegetarian    & Overall       \\ \midrule
Rule       & \textit{0.00}          & \textit{0.00}           & \textit{0.00}         & \textit{0.00}         \\ \midrule
MT         & 6.85           & 22.37          & 0.49          & 6.50          \\
+Filter    & 6.85           & 22.37          & 0.49          & 6.50          \\ \midrule
E2E        & 13.70          & 10.53          & 0.00          & 5.08          \\
+Filter    & 13.70          & 10.53          & 0.00          & 5.08          \\ \midrule
E2E-P      & 6.85           & 14.47          & 0.98          & 5.08          \\
+Filter    & 6.85           & 14.47          & 0.98          & 5.08          \\ \midrule
SHARE      & 6.85           & 14.47          & 0.00          & 4.52          \\
+Filter    & \textbf{4.11}           & \textbf{11.84}          & 0.00          & \textbf{3.39}          \\ \bottomrule
\end{tabular}
\caption{Percent of edited recipe directions referencing ingredients that violate target hard dietary restrictions.}
\label{tab:cat-err-step}
\end{table}

\section{Human Studies and Discussion}
\label{human}

While automatic metrics can help gauge recipe appropriateness, we need human studies to accurately determine whether a recipe---a series of actions performed on a set of ingredients---can be followed.

\paragraph{Human Evaluation}
We randomly sampled eight recipes for each constraint, with each crowd-sourced judge---home cooks familiar with our dietary constraints---given the ground truth and versions edited by each model (total 336 recipes);
each recipe was reviewed by two different judges to determine if it violated the target dietary constraint along with providing additional written feedback.
Evaluators were presented with a single recipe at a time in random order and were not told the source of each recipe.

We report strong inter-annotator agreement for constraint violation, with a Kendall's Tau \cite{kendall1938new} value of 0.762.
We find a significant difference in the constraint violation rate between models ($p=0.013$) via a $\chi^2$ test, with SHARE generating the most appropriate outputs across constraints.
Some evaluators noted violations of strict constraints, but further inspection revealed these to be safe (e.g.~eggs do not trigger dairy allergies).
Soft constraint satisfaction remains subjective:
a low-carb diet may allow all types of carbs in moderation or disallow specific types (e.g.~complex starches).
Thus, even some ground truth and Rule baseline recipes can violate a target soft constraint.

77 judges provided extra written feedback, and we used qualitative thematic coding \cite{gibbs2007thematic} to identify common themes (\Cref{tab:human-eval}).
In 13\% of recipe directions, judges noticed references to ingredients not mentioned in the ingredients list (Unlisted ingredient).
In ground truth recipes, this can comprise optional flavorings (e.g.~salt)
and garnishes (e.g.~fruits),
while LM baselines
often reference unlisted, substantive ingredients (e.g.~vegetables and starches).
SHARE consistently produced recipes with fewer unlisted references compared to baselines.

Other feedback reflected how stylistic preferences affect perceptions of recipe quality.
Several judges felt it confusing when an ingredient was referred to by a synonym or general phrase (Inconsistent naming: e.g.~\emph{cheddar} referred to as \emph{cheese}, or mentions of both \emph{cilantro} and \emph{coriander}).
In 5\% of recipes (including ground truth), judges asked for more detail in the instructions or ingredients (Lacking detail).
Some judges wondered whether listed ingredients were pre-processed or raw, and others wanted to know specific utensils and dimensions for manipulating (e.g.~folding) or cutting ingredients.

\begin{table}[t!]
\centering
\small
\begin{tabular}{@{}lrrrrrr@{}}
\toprule
Issue      &  Gold & Rule & MT   & E2E  & E2E-P & \textbf{Ours} \\ \midrule
Constraint & 10.7 & 15.2 & 14.3 & 24.1 & 8.9   & \textbf{8.0}  \\
Unlisted   & {\ul 7.1}  & 22.3 & 14.3 & 12.5 & 14.3  & \textbf{7.1}  \\
Naming     & 4.2  & 2.1  & \textbf{0.0}  & 2.1  & 10.4  & 6.3  \\
Detail (S) & {\ul 4.2}  & 14.6 & 6.3  & 8.3  & 8.3   & \textbf{6.3}  \\
Detail (I) & 2.1  & \textbf{0.0}  & 4.2  & 4.2  & 6.3   & 2.1  \\ \bottomrule
\end{tabular}
\caption{Issues in edited recipes identified by human evaluators (\%): constraint violation, unlisted ingredients, inconsistent naming, lacking step details, lacking ingredient details.}
\label{tab:human-eval}
\end{table}

\paragraph{Cooking Study}

Ultimately, one can best judge recipe quality by attempting to cook the dish.
We recruited seven home cooks with 3+ years of cooking experience, each tasked to cook 3 random recipes edited by SHARE (total of 21 different recipes covering all seven dietary constraints).
Cooks were instructed to follow recipes exactly, recording 1) how complete the recipe was (i.e.~missing ingredients/steps); 2) how difficult it was to infer ingredient amounts; 3) overall difficulty of execution; and 4) if the recipe was appropriate for its target category.
\Cref{tab:qual-ex-dijon} shows an example recipe.

No cooks reported difficulties following any recipe.
In 90\% of the cases, cooks reported using `common sense' or `cooking experience' to `easily' infer ingredient amounts.
All cooks agreed that the recipes received were appropriate for their dietary constraint.
The most common (19\%) complaint among cooks was that the final dish did not resemble the base recipe name.
This suggests that there is room for our models to better learn specific textures, shapes, and techniques associated with different food types---while SHARE edits recipes in constraint-respecting and diverse ways, they may surprise user in the process (e.g.~editing \textit{curries} into \textit{stews}).
Ultimately, we demonstrate that SHARE can successfully create recipes that not only satisfy a specified dietary constraint but are delicious and easily followed by home cooks.

\begin{table}[t!]
\small
\centering
\begin{tabular}{@{}ll@{}}
\toprule
\multicolumn{1}{c}{\textbf{Base (Chicken Dijon)}} &
  \multicolumn{1}{c}{\textbf{SHARE (Low-Calorie)}} \\ \midrule
\begin{tabular}[c]{@{}l@{}}Butter, Chicken, Chicken\\broth, Dijon mustard,\\Flour, Light cream, Onion,\\Pepper, Salt, Wine\end{tabular} &
  \begin{tabular}[c]{@{}l@{}}Chicken, Dijon mustard,\\Flour, Pepper, Salt, \emph{\textcolor{blue}{Carrot}},\\\emph{\textcolor{blue}{Garlic, Mushroom, Soy}}\\\emph{\textcolor{blue}{sauce.}} \textcolor{red}{\sout{Butter, Chicken}}\\\textcolor{red}{\sout{broth, Light cream, Onion,}}\\ \textcolor{red}{\sout{Wine}}\end{tabular} \\ \midrule
\begin{tabular}[c]{@{}l@{}}1) Sprinkle chicken breasts\\with onion powder, lemon\\pepper, salt.\\ 2) Saute the chicken in the\\butter for 20 min, or cooked\\through.\\ 3) Remove chicken to a\\platter   and keep warm.\\ 4) Measure pan juices and\\   chicken broth for 1 cup.\\ 5) Return the stock mixture\\to the pan and add wine.\\ 6) Stir together the light\\ cream and flour till smooth.\\ 7) Add to the broth, cook,\\stir till smooth, thickened\\and bubbly.\\ 8) Stir in the dijon mustard,\\   stir till smooth.\\ 9) Return the chicken to pan\\or serve the sauce separately.\end{tabular} &
  \begin{tabular}[c]{@{}l@{}}1) Preheat oven to 350 f.\\ 2) Place chicken breasts,\\breast side up, in a shallow\\roasting pan.\\ 3) Combine \textcolor{blue}{\emph{soy sauce}},\\mustard, \textcolor{blue}{\emph{garlic}} and salt.\\ 4) Brush over both sides\\of the chicken breasts.\\ 5) Bake for 40 minutes.\\ 6) Meanwhile, heat a\\small sauce-pan over\\medium heat.\\ 7) Stir in \textcolor{blue}{\emph{carrots}}, pepper,\\   and \textcolor{blue}{\emph{mushrooms}}.\\ 8) Cook and stir 5 minutes.\\ 9) Stir in flour.\\ 10) Pour over chicken\\and bake for extra 10 mi-\\nutes until cooked through.\\ 11) Serve chicken with\\carrots and sauce.\end{tabular} \\ \bottomrule
\end{tabular}
\caption{SHARE creates a low-calorie version of Chicken Dijon (left: original).
\emph{\textcolor{blue}{Italics}} and \textcolor{red}{\sout{strike through}} denote added and removed ingredients respectively.}
\label{tab:qual-ex-dijon}
\end{table}

\subsection{Remaining Challenges}
As our step predictor depends solely on input ingredients, it can mix cooking techniques to change the form of the edited dish (e.g.~casserole in place of pasta or stew in place of curry).
While we experimented with including the original recipe name as an input during our step generation process, in preliminary experiments we found that including the name increased the rate of copying unsatisfactory (i.e. using original ingredients or text that violated the target dietary constraint) passages from the base recipe, similar to baselines.

To better serve cooks who crave specific dishes, we hope to extend our work in the future by learning structured representations of recipes \cite{DBLP:conf/iclr/BosselutLHEFC18} 
and generating recipe instructions as procedural directed graphs \cite{DBLP:conf/lrec/MoriMYS14}
instead of raw text.

Cooks from different cultural contexts often exhibit different patterns of preferences and requirements in their desired recipes \cite{DBLP:conf/icwsm/ZhangTLE19}, and in the future we hope to extend our work to controllable recipe editing using more subtle traits such as cuisines (e.g.~making a Chinese version of meatloaf).

\section{Conclusion}
In this work, we propose the task of controllable recipe editing and present a System for Hierarchical Assistive Recipe Editing (SHARE) to tackle this task in the context of dietary constraints.
On a novel dataset of 83K pairs of recipes and
versions that satisfy dietary constraints, 
we demonstrate that this is a challenging task that cannot be solved with rule-based ingredient substitutions or straightforward application of existing recipe generators.
We show that SHARE can edit recipes to produce coherent, diverse versions that accommodate desired dietary restrictions.
Human evaluation and cooking trials also reveal that SHARE produces feasible recipes that can be followed by home cooks to create appealing dishes.
In the future, we aim to leverage recipe workflows to generate flow graphs \cite{DBLP:conf/lrec/MoriMYS14} and explore editing recipes to accommodate more complex preferences like cuisines.

\section*{Limitations}

We have demonstrated that SHARE can edit recipes in ways that are more human-like and appropriate compared to human-written substitution rules or state-of-the-art recipe generation models.
In this section, we highlight several limitations and continued challenges facing our SHARE approach and the task of assistive recipe editing.

\paragraph{Recipe Coherence}
While recipes can be regarded
as structured instruction sets, they are written as natural text.
Neither SHARE nor other language generation approaches for recipe generation can guarantee that their output will produce edible food or that directions make sense in context and in order.
\citet{DBLP:conf/iclr/BosselutLHEFC18} explore training models to understand procedural text by predicting state changes between mentioned ingredients, but report substantial room for improvement in accuracy for generating directions.
In human evaluation we find that recipes generated by SHARE are readable and understandable by human judges and home cooks, but there remain opportunities to improve coherence.
In future work we hope to frame recipe generation as a structured graph generation problem to enable coherence constraints (e.g.~flour and water must be combined to form dough before it can be rolled) during the generation process.

\paragraph{Recipe Safety}
Similar to coherence, the natural-text modality of our recipe inputs and outputs raises the possibility of unsafe ingredients being written into the recipe directions (e.g.~butter being mentioned in a dairy-free recipe).
We demonstrate in Tables 4 and 5 that SHARE creates safer, more appropriate recipes than other generation methods, but a small risk remains.
As such, we caution against naive usage of recipe editing models (including SHARE), and stress the importance of manual inspection and review of edited recipes---either by the home cooks themselves or a team of reviewers if such a model is used as part of a service.

\paragraph{Ingredient Quantities}
While SHARE can predict which ingredients to use and how to make use of them, like previous work in recipe generation \cite{DBLP:conf/www/LeeKAPLLV20,DBLP:conf/emnlp/MajumderLNM19,marin2019learning,DBLP:conf/iclr/BosselutLHEFC18,DBLP:conf/emnlp/KiddonZC16} it does not output amounts or units.
We found in human evaluations that judges could read and understand recipes without units or quantities given, and home cooks with some experience could easily infer ingredient quantities.
However, predicting ingredient quantities remains an important challenge to make assistive technologies more accessible for beginner cooks, as mis-inferring ingredient ratios (e.g.~sugar and butter) can have negative taste and textural impacts on cooking results \cite{HyoGee2004SensoryAM}.
In future work we will explore the ratio of pairs of ingredients within a recipe.

\bibliography{tacl2018}

\newpage
\appendix

\section{Additional Experimental Details}
The SHARE ingredient model comprises a 6-layer Transformer encoder-decoder with a hidden size of 128.
The step generator comprises an 8-layer Transformer encoder-decoder with a hidden size of 256 and a 64-dimensional factorized token embedding \cite{DBLP:conf/iclr/LanCGGSS20}.
We train models using the LAMB optimizer \cite{DBLP:conf/iclr/YouLRHKBSDKH20}, fine-tuning baseline models for 10 epochs and training SHARE for up to 100 epochs.
We generate recipe steps via greedy sampling.
We report results for the median of three experiments.
Model training and fine-tuning hyperparameters was determined through a small-scale grid search for learning rate $[1e-2, 1e-3, 1e-4]$.

\section{Dataset and Code}
We use the Food.com recipe data only for academic non-commercial research purposes and the data remains under copyright of the original authors.
Food.com is a primarily United States-based English recipe aggregator website.
We do not track user demographics.

We do not use user-names, user IDs, or other personally identifiable information.
We only make use of recipe ingredient lists, instruction sets, and dietary restriction tags are used to train SHARE.
All statistics reported are in aggregate; we do not and will not store or release any user activity data as part of this paper.

We will release the RecipePairs dataset and code for SHARE under the MIT license\footnote{\url{https://github.com/shuyangli94/share-recipe-editing}}.

\section{Human Evaluation and Study Details}
\paragraph{Human Evaluation}
We recruited evaluators using the Amazon Mechanical Turk platform, restricting to native English speakers with a 99\% acceptance rate.\footnote{\url{https://www.mturk.com/}}
We included a message on the evaluation page indicating the purpose of the task was for a research study.
The platform affords workers an option to reject the task.
Judges were home cooks familiar with our dietary constraints: on average, they reported cooking 6.3 meals per week, with 68.3\% of them `sometimes', `often', or `always' eating food satisfying the target dietary constraint.
We did not track judge demographics other than dietary constraint familiarity and cooking experience.
Evaluators were asked the question ``Does this recipe use ingredients that violate the dietary constraint?'' and further asked to provide free-form textual feedback if any.

\begin{figure}[t!]
  \includegraphics[width=\columnwidth]{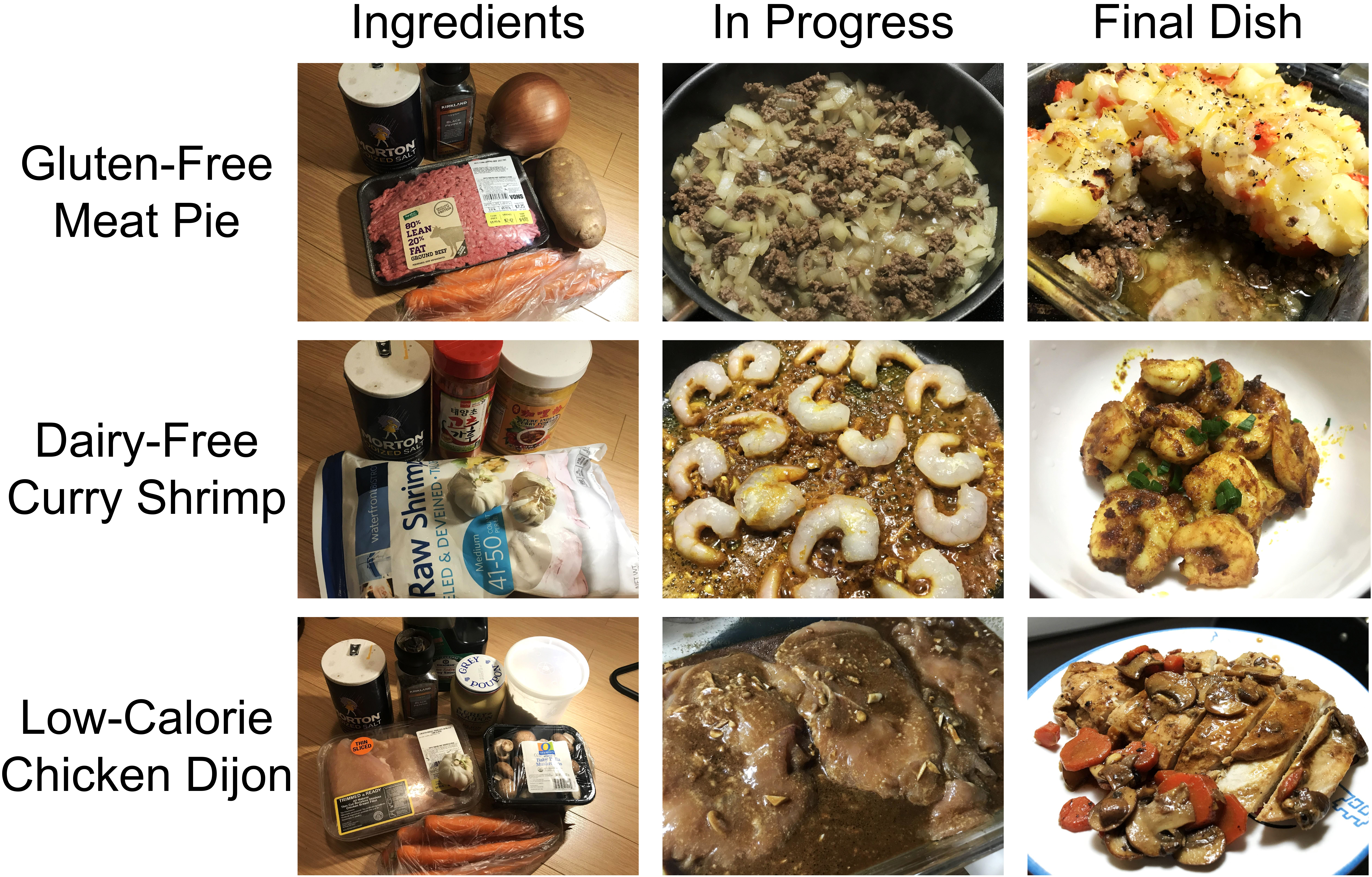}
  \caption{Cooked dishes following our model-edited recipes.}
  \label{fig:results-food}
\end{figure}

\paragraph{Cooking Study}
Home cooks were recruited from university student populations in the United States, with the requirement of fluent or native English writing and reading comprehension skills.
Cooks were informed that the purpose of the cooking study was for academic research, and given an opportunity to decline to cook any of the recipes assigned---no cooks chose to decline.
Cooks were asked for dietary restrictions and food allergies, and instructed to stop cooking and provide feedback without further action if they felt any health-related risks.
Two cooks were vegetarian, and two other cooks were dairy-intolerant (lactose intolerance).
Several cooks took pictures of the cooking process, with three such sets of photos displayed in \Cref{fig:results-food}.

\section{Additional Recipe Examples}
\label{appendix_examples}

We present several additional representative samples of recipes edited using the \textbf{E2E-P}, \textbf{MT}, and \textbf{SHARE} models.

In \Cref{tab:qual-ex-squash}, our models were asked to produce a low-fat version of Squash Casserole.
All three recipes were able to identify high-fat items from the base recipe (stuffing mix, cream of chicken soup, margarine, sour cream) and remove them.
Here we see that human-written recipes can also incorporate ingredients not found in the original ingredients list (e.g.~cheese to top the casserole).

In \Cref{tab:qual-ex-tofu}, our models produced dairy-free versions of Tofu Tikka Masala, a dish of tofu in a creamy curry sauce.
We observe that large language models like those used for E2E-P and MT can often reference extraneous ingredients because they match common phrases or patterns in general cooking language---the MT model does not include yogurt in the ingredients list but still uses it in the recipe directions.
Interestingly, the hierarchical model eschews tofu in favor of shrimp, making a dairy-free creamy shrimp curry.

\begin{table*}[t!]
\small
\centering
\caption{An example of controllable recipe editing. We ask our E2E-P, MT, and SHARE models to create a low-fat version of Squash Casserole (base recipe on left). \emph{\textcolor{blue}{Italics}} and \textcolor{red}{\sout{strike through}} denote added and removed ingredients respectively.}
\label{tab:qual-ex-squash}
\begin{tabular}{@{}llll@{}}
\toprule
\multicolumn{1}{c}{\textbf{Base (Squash Casserole)}} &
  \multicolumn{1}{c}{\textbf{E2E-P (Low-Fat)}} &
  \multicolumn{1}{c}{\textbf{MT (Low-Fat)}} &
  \multicolumn{1}{c}{\textbf{SHARE (Low-Fat)}} \\ \midrule
\begin{tabular}[c]{@{}l@{}}Cream of chicken soup, Egg,\\ Herb stuffing mix, Margarine,\\ Onion, Sour cream, Squash\end{tabular} &
  \begin{tabular}[c]{@{}l@{}}Onion, Squash, \emph{\textcolor{blue}{Basil}},\\ \emph{\textcolor{blue}{Garlic}}, \emph{\textcolor{blue}{Mushroom}},\\ \emph{\textcolor{blue}{Parmesan}}, \emph{\textcolor{blue}{Pepper}}, \emph{\textcolor{blue}{Salt}}\\ \emph{\textcolor{blue}{Tomato}}, \emph{\textcolor{blue}{Zucchini}}\\ \textcolor{red}{\sout{Cream of chicken soup}},\\ \textcolor{red}{\sout{Egg}} \textcolor{red}{\sout{Herb stuffing mix}},\\ \textcolor{red}{\sout{Margarine}}, \textcolor{red}{\sout{Sour Cream}}\end{tabular} &
  \begin{tabular}[c]{@{}l@{}}Egg, Onion, Squash,\\ \emph{\textcolor{blue}{Breadcrumb}}, \emph{\textcolor{blue}{Butter}}, \emph{\textcolor{blue}{Cheddar}},\\ \emph{\textcolor{blue}{Milk}}, \emph{\textcolor{blue}{Pepper}}, \emph{\textcolor{blue}{Salt}}\\ \textcolor{red}{\sout{Cream of chicken soup}},\\ \textcolor{red}{\sout{Herb stuffing mix}}, \textcolor{red}{\sout{Margarine}},\\ \textcolor{red}{\sout{Sour cream}}\end{tabular} &
  \begin{tabular}[c]{@{}l@{}}Egg, Onion, Squash,\\ \emph{\textcolor{blue}{Breadcrumb}}, \emph{\textcolor{blue}{Cheddar}},\\ \emph{\textcolor{blue}{Pepper}}, \emph{\textcolor{blue}{Salt}}, \emph{\textcolor{blue}{Zucchini}}\\ \textcolor{red}{\sout{Cream of chicken soup}},\\ \textcolor{red}{\sout{Herb stuffing mix}},\\ \textcolor{red}{\sout{Margarine}}, \textcolor{red}{\sout{Sour cream}}\end{tabular} \\ \midrule
\begin{tabular}[c]{@{}l@{}}1) Combine soup, sour cream\\   and eggs.\\ 2) Fold into squash and onion.\\ 3) Melt butter and pour over\\   stuffing mix.\\ 4) Line bottom of 9x13-inch\\   pyrex bowl with 1/2 of\\   stuffing mix.\\ 5) Pour squash mix in bowl and\\   top with other 1/2 of stuffing.\\ 6) Bake at 350° for 30 to\\   40 minutes.\\ 7) Turn oven off and sprinkle\\   with grated american cheese\\   and leave in oven to melt.\end{tabular} &
  \begin{tabular}[c]{@{}l@{}}1) Preheat oven to \\350 degrees.\\ 2) Spray a 2 quart\\ casserole dish with\\nonstick cooking spray.\\ 3) In a large bowl,\\ combine all\\   ingredients except \\parmesan cheese.\\ 4) Pour into prepared\\ casserole dish.\\ 5) Sprinkle with \\parmesan cheese.\\ 6) Bake for 30 minutes.\end{tabular} &
  \begin{tabular}[c]{@{}l@{}}1) Preheat oven to 350 degrees.\\ 2) Cook squash in boiling water\\   until tender.\\ 3) Drain well.\\ 4) Mash squash.\\ 5) Add butter, milk, salt, pepper,\\   onion, egg, and 1/2 cup of cheese.\\ 6) Mix well.\\ 7) Pour into a greased 2 quart\\   casserole dish.\\ 8) Combine bread crumbs,\\   remaining 1/2 cup cheese, and\\   melted butter.\\ 9) Sprinkle over squash mixture.\\ 10) Bake for 30 minutes.\end{tabular} &
  \begin{tabular}[c]{@{}l@{}}1) Preheat oven to\\ 350 degrees.\\ 2) Combine ingredients\\   except for cheese.\\ 3) Pour mixture into \\greased   casserole dish.\\ 4) Bake for 40 minutes or\\ until lightly browned.\\ 5) Top with cheese, return\\ to oven for a minute.\end{tabular} \\ \bottomrule

\end{tabular}
\end{table*}

\begin{table*}[ht!]
\centering
\small
\caption{An example of controllable recipe editing. We ask our E2E-P, MT, and SHARE models to create a dairy-free version of Tofu Tikka Masala (base recipe on left). \emph{\textcolor{blue}{Italics}} and \textcolor{red}{\sout{strike through}} denote added and removed ingredients respectively.}
\label{tab:qual-ex-tofu}
\begin{tabular}{@{}llll@{}}
\toprule
\multicolumn{1}{c}{\textbf{Base (Tofu Tikka Masala)}} &
  \multicolumn{1}{c}{\textbf{E2E-P (Dairy-Free)}} &
  \multicolumn{1}{c}{\textbf{MT (Dairy-Free)}} &
  \multicolumn{1}{c}{\textbf{SHARE (Dairy-Free)}} \\ \midrule
\begin{tabular}[c]{@{}l@{}}Chili powder, Coriander,\\ Garlic, Ginger, Jalapeno\\ chile, Lemon juice, Onion,\\ Salt, Soymilk, Turmeric,\\ Vegetable oil, Yogurt\end{tabular} &
  \begin{tabular}[c]{@{}l@{}}Coriander, Garlic, Ginger,\\ Onion, Salt, Turmeric,\\ Vegetable oil, \emph{\textcolor{blue}{Bean}},\\ \emph{\textcolor{blue}{Cinnamon}}, \emph{\textcolor{blue}{Cumin}},\\\emph{\textcolor{blue}{Pepper}}, \emph{\textcolor{blue}{Tomato}}, \emph{\textcolor{blue}{Water}}\\ \textcolor{red}{\sout{Chili powder}}, \textcolor{red}{\sout{Jalapeno chile}},\\ \textcolor{red}{\sout{Lemon juice}}, \textcolor{red}{\sout{Soymilk}},\\\textcolor{red}{\sout{Yogurt}}\end{tabular} &
  \begin{tabular}[c]{@{}l@{}}Chili powder, Coriander,\\ Garlic, Ginger, Lemon \\juice, Onion, Salt,\\ Turmeric, Vegetable oil,\\ \emph{\textcolor{blue}{Cumin}}, \emph{\textcolor{blue}{Pepper}}, \emph{\textcolor{blue}{Tofu}},\\ \emph{\textcolor{blue}{Water}}  \textcolor{red}{\sout{Jalapeno chile}},\\ \textcolor{red}{\sout{Soymilk}}, \textcolor{red}{\sout{Yogurt}}\end{tabular} &
  \begin{tabular}[c]{@{}l@{}}Coriander, Garlic, Ginger,\\ Onion, Salt, Turmeric,\\ Vegetable oil, \emph{\textcolor{blue}{Coconut}} \emph{\textcolor{blue}{milk}},\\ \emph{\textcolor{blue}{Cumin}}, \emph{\textcolor{blue}{Curry}}, \emph{\textcolor{blue}{Masala}}, \emph{\textcolor{blue}{Pea}},\\ \emph{\textcolor{blue}{Pepper}}, \emph{\textcolor{blue}{Rice}}, \emph{\textcolor{blue}{Shrimp}},\\ \emph{\textcolor{blue}{Tomato}}, \emph{\textcolor{blue}{Water}} \textcolor{red}{\sout{Chili powder}},\\ \textcolor{red}{\sout{Jalapeno chile}}, \textcolor{red}{\sout{Lemon juice}},\\ \textcolor{red}{\sout{Soymilk}}, \textcolor{red}{\sout{Yogurt}}\end{tabular} \\ \midrule
\begin{tabular}[c]{@{}l@{}}1) Mix together lemon\\ juice, ginger, garlic, green\\ chili, half of the cilantro,\\ chili powder, salt, and tofu.\\ 2) Cover and leave to\\ marinate for an hour.\\ 3) Heat oil in a pan and\\ cook onions for 5 minutes\\ on medium heat until\\ translucent.\\ 4) Add turmeric powder,\\   yogurt, soymilk and the\\   remaining cilantro.\\ 5) Stir for a few minutes\\ until sauce thickens.\\ 6) Add the tofu mixture\\ into the sauce and cook\\ for 5 minutes on medium.\\ 7) Adjust seasoning.\\ 8) Serve garnished with\\   fresh cilantro with your\\   choice of indian nan\\ bread  or heated rice.\end{tabular} &
  \begin{tabular}[c]{@{}l@{}}1) Heat oil in large skillet\\ over medium heat.\\ 2) Add onion and cook\\ until softened, about\\5 minutes.\\ 3) Add garlic and ginger \\and cook for 1 minute.\\ 4) Add tomatoes, water,\\ cumin, coriander, turmeric,\\ cinnamon, and cayenne\\ and bring to a boil.\\ 5) Reduce heat and simmer \\for 10 minutes.\\ 6) Add tofu and simmer\\for 10 minutes.\\ 7) Season with salt and\\ pepper.\end{tabular} &
  \begin{tabular}[c]{@{}l@{}}1) To make the marinade,\\   combine the lemon juice,\\ water, oil, garlic, ginger,\\ cumin, coriander, turmeric,\\ salt, and pepper in a bowl.\\ 2) Add the tofu and toss\\ to coat.\\ 3) Cover and refrigerate\\ for at least 1 hour or up to\\ overnight.\\ 4) For sauce, combine\\ yogurt, cilantro, green\\ onions and garlic.\\ 5) Cover and refrigerate.\\ 6) Preheat the oven to 350.\\ 7) Heat the oil in a large \\skillet over medium-high\\ heat.\\ 8) Add the tofu and cook,\\   stirring occasionally, until\\   browned, about 5 minutes.\\ 9) Transfer to a baking \\sheet and bake until the\\ tofu is cooked through, \\10 minutes.\\ 10) To serve, spoon tofu\\   onto plates and top\\ with sauce.\end{tabular} &
  \begin{tabular}[c]{@{}l@{}}1) Heat the oil in a skillet.\\ 2) Stir-fry cumin until\\ fragrant.\\ 3) Add the onions and\\ saute until they start\\ to turn translucent.\\  4) Add in the turmeric,\\ garam masala and cayenne \\and saute until fragrant,\\ about 1 minute.\\ 5) Stir in the curry powder.\\ 6) Stir in the coconut milk.\\ 7) Add the water and\\ tomatoes,   bring to a boil,\\ then simmer.\\ 8) Cover and cook, stirring\\   occasionally, until the rice\\ is tender, about 30 minutes.\\ 9) Stir in the peas, salt, and\\   pepper, then cover the pan.\\ 10) Simmer until the peas\\ are just tender, about\\ 5 minutes.\\ 11) Stir in the shrimp and\\   simmer for 3 to 4 minutes.\\ 12) Stir in the chopped\\ cilantro, and season with\\ salt and pepper.\end{tabular} \\ \bottomrule
\end{tabular}
\end{table*}

\end{document}